\documentclass[runningheads]{llncs}

\usepackage{eccv}

\usepackage{eccvabbrv}

\usepackage{graphicx}
\usepackage{booktabs}
\usepackage[linesnumbered,ruled]{algorithm2e}
\usepackage{colortbl}
\usepackage{multirow}
\usepackage{amsmath}
\usepackage{marvosym}
\usepackage{bm}

\usepackage[accsupp]{axessibility}  

\usepackage{hyperref}

\usepackage{orcidlink}

\begin{document}

\title{Straightforward Layer-wise Pruning for More Efficient Visual Adaptation} 

\titlerunning{Straightforward Layer-wise Pruning for More Efficient Visual Adaptation}

\author{Ruizi Han \and
Jinglei Tang\textsuperscript{(\Letter)}
}

\authorrunning{R.~Han et al.}

\institute{College of Information Engineering, Northwest A\&F University, Yangling, China
\email{rzh@nwafu.edu.cn}, \email{tangjinglei@nwsuaf.edu.cn}}

\maketitle

\begin{abstract}
Parameter-efficient transfer learning (PETL) aims to adapt large pre-trained models using limited parameters. While most PETL approaches update the added parameters and freeze pre-trained weights during training, the minimal impact of task-specific deep layers on cross-domain data poses a challenge as PETL cannot modify them, resulting in redundant model structures. Structural pruning effectively reduces model redundancy; however, common pruning methods often lead to an excessive increase in stored parameters due to varying pruning structures based on pruning rates and data. Recognizing the storage parameter volume issue, we propose a Straightforward layer-wise pruning method, called SLS, for pruning PETL-transferred models. By evaluating parameters from a feature perspective of each layer and utilizing clustering metrics to assess current parameters based on clustering phenomena in low-dimensional space obtained through t-SNE, SLS facilitates informed pruning decisions. Our study reveals that layer-wise pruning, with a focus on storing pruning indices, addresses storage volume concerns. Notably, mainstream Layer-wise pruning methods may not be suitable for assessing layer importance in PETL-transferred models, where the majority of parameters are pre-trained and have limited relevance to downstream datasets. Comparative analysis against state-of-the-art PETL methods demonstrates that the pruned model achieved a notable balance between model throughput and accuracy. Moreover, SLS effectively reduces storage overhead arising from varying pruned structures while enhancing the accuracy and speed of pruned models compared to conventional pruning methods. The code is available at \href{https://github.com/RuiZiHan/SLS}{https://github.com/RuiZiHan/SLS.}
\end{abstract}

\section{Introduction}
\label{sec:intro}
\begin{figure*}[t]
    \centering
\includegraphics[width=1\textwidth]{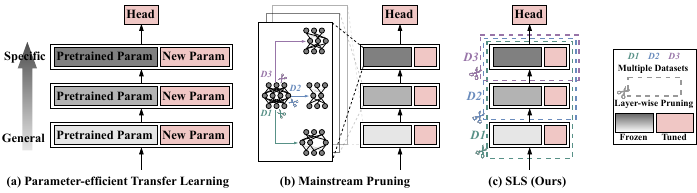}
    \caption{Comparison of Parameter-efficient Transfer Learning, Mainstream Pruning, and our SLS. SLS can selectively prune task-specific layers that cannot be updated by PETL, without increasing the number of stored parameters.}
    \label{fig1}
\end{figure*}

Parameter-efficient transfer learning (PETL) has garnered significant attention for its capacity to adapt large pre-trained models with limited parameters\cite{r1}, making it a valuable tool for AI model users and researchers facing hardware constraints. PETL involves tuning a pre-trained model by adding or selecting a small number of parameters while freezing the pre-trained weights during data transfer, updating only the selected or newly added parameters. While PETL plays a crucial role in generative\cite{r2} and predictive\cite{2019firstnlp} models, the inability to adjust frozen parameters, such as task-specific parameters in the top layer, as shown in \cref{fig1}(a), can lead to redundant model structures, ultimately consuming computing resources without commensurate accuracy gains.

For the redundant parameters in the model, a common practice is to reduce the redundant parameter volume through structural pruning methods. For example, based on magnitude \cite{mg_pruning_1,mg_pruning_2,mg_pruning_3} and based on gradient \cite{gr_pruning_1,gr_pruning_2}. However, despite the usefulness of these methods for conventional training models, through our research, we identified two unavoidable issues when applied to models after PETL transfer: 
\begin{itemize}
\setlength{\itemsep}{4pt}
    \item It significantly increases the model's storage overhead. As shown in \cref{fig1}(b), due to different downstream data, the pruning rate set by the pruning method varies, leading to different network structures for different datasets, resulting in a large storage space occupied by these diverse structures.
    \item It leads to a loss in model accuracy. For models after PETL transfer, the pruned redundant parameters are from the pre-trained parameters in the network, which are not directly related to the downstream dataset. The trainable parameters in the model need to adapt to the pruned new structure during the retraining process, which our experimental results suggest hinders the model's accuracy recovery.
\end{itemize}

In response to these challenges, we introduce the \textbf{S}traightforward \textbf{L}ayer-wi\textbf{S}e Pruning method (SLS) tailored for PETL-transferred models, as shown in \cref{fig1}(c). SLS operates at the layer level, targeting Task-Specific parameters and the pruned layer index for diverse downstream data, thus maintaining the low storage parameter quantity characteristic of PETL models. By pruning at the layer granularity, traditional magnitude-based and gradient-based parameter evaluation criteria are challenging to apply. Therefore, leveraging feature outputs from different layers, we employ dimensionality reduction using t-SNE\cite{t-SNE} and evaluate features through the Silhouette Coefficient Index (SC\_Index)\cite{SCIndex}, dynamically determining pruning layer numbers for varying downstream datasets.

To validate SLS, experiments were conducted on PETL-transferred models using Convpass\cite{convpass} and RepAdapter\cite{repadapter} methods, including image classification tasks with Vision Transformers (ViT)\cite{vit} and Swin Transformers (Swin)\cite{swin}, object detection tasks with RetinaNet\cite{retinanet} and image segmentation tasks with SETR-PUP\cite{setr}. The experimental results show that SLS significantly improves model throughput with minimal loss of accuracy while maintaining the low storage parameter quantity of PETL methods and without introducing additional training. Compared to the mainstream pruning method DepGraph\cite{depgraph}, SLS outperforms in model accuracy, inference throughput, and storage parameter quantity. 

In summary, our contributions can be summarized as follows:
\begin{itemize}
\setlength{\itemsep}{4pt}
\item We propose a pruning method SLS for PETL models, demonstrating the presence of a large number of redundant parameters in models after PETL transfer when there is a significant gap between downstream and pre-trained datasets.
\item We propose an intuitive feature-level analysis method, providing a new perspective for evaluating the importance of structural pruning parameters.
\item SLS surpasses the current mainstream structural pruning method DepGraph in model storage, accuracy, and speed on the VTAB-1k\cite{vtab} benchmark with the same pruning parameter quantity, using a simple strategy.
\end{itemize}

\section{Related Work}
\subsection{Parameter-efficient Transfer Learning}
Parameter-efficient transfer learning (PETL) was first proposed in natural language processing (NLP)\cite{2019firstnlp,lora,petl_nlp_1,petl_nlp_2,petl_nlp_3,petl_nlp_4,petl_nlp_5} and has garnered significant attention due to its capacity to use a small number of parameters to achieve or surpass the results of full fine-tuning large-scale pre-trained models. PETL methods aim to select or add a limited number of parameters to the pre-trained model and tune those parameters during training. The objective is to adapt the pre-trained model with a minimal number of parameters. Recently, an increasing amount of computer vision (CV) research has been employing this technology, specifically add-based\cite{convpass,adaptformer}, prompt-based\cite{vpt,pro-tuning}, and reparameterization-based\cite{repadapter,lora,ssf} methods. Add-based methods, for example, AdaptFormer\cite{adaptformer}, introduce learnable Adapter modules to the pre-trained model either in parallel or serially. Prompt-based approaches, like VPT\cite{vpt}, incorporate learnable prompt tokens into the input section of the model. Reparameterization-based approaches, such as LoRA\cite{lora}, learn low-rank parameters for the frozen weights of multi-head attention and merge the learned parameters into the pre-trained model during inference.

\subsection{Structural Pruning}
Structural pruning is a common and effective model compression method, which usually evaluates the importance of various structures in the pre-trained model by some predefined evaluation strategy, and prunes the structure according to the evaluation result to achieve the purpose of model acceleration. Due to its hardware-friendly nature, it can improve the speed of the model without complex construction of special acceleration algorithms\cite{pruning_speed_up}, thus has been widely used\cite{pruning_wide_use_1,pruning_wide_use_2}. Structural pruning algorithm design can be mainly categorized into magnitude-based\cite{mg_pruning_1,mg_pruning_2,mg_pruning_3}, gradient-based\cite{gr_pruning_1,gr_pruning_2} and mask-based\cite{vitslim}. Magnitude-based methods assume that the importance of a parameter can be quantified by its magnitude, e.g., \cite{mg_pruning_1} uses the paradigm of model structure to measure its importance. Gradient-based methods assume that the importance of a parameter is related to the loss value obtained from its input data, e.g., \cite{hessian_pruning} uses the hessian trace as the sensitivity metric for the pruning model. Mask-based pruning uses differentiable masks to indicate the global importance of dimensions in various modules. These masks start with values of 1 and are trained to become sparse, allowing for the determination of the relative importance of each dimension. 

Layer-wise pruning is a branch of structural pruning. LayerPrune\cite{layer_prune_magnitude} evaluates the importance of the entire layer structure based on the magnitude and gradient information of the parameters within the layer, which is feasible for models trained conventionally. However, for models after PETL transfer, most of the information about the parameters within the layer comes from pre-trained weights, which do not reflect the importance for downstream data. Furthermore, as shown in \cref{fig2}, solely using the magnitude information of the newly added parameters still fails to reflect the importance of the corresponding layer. Chen\cite{layer_prune_classifier} trains separate classifiers for each layer's intermediate features to determine the importance of the layer. For small-scale models and datasets, this method can accurately assess the importance of the layers. However, as the model and dataset scale increase, training for each layer incurs significant additional overhead that cannot be ignored. 

Our SLS does not require training an additional classification head. Instead, it utilizes the unsupervised feature dimension reduction method t-SNE\cite{t-SNE} for feature selection and directly evaluates features using clustering evaluation metric SC\_Index\cite{SCIndex}, thereby avoiding significant additional overhead.

\section{Method}
\subsection{Preliminary}
We begin by revisiting the commonly used dimensionality reduction algorithm, t-distributed stochastic neighborhood embedding (t-SNE)\cite{t-SNE}, as well as the clustering algorithm metric, silhouette coefficient index (SC\_Index)\cite{SCIndex}.
\begin{figure}[t]
    \centering
\includegraphics[width=1\textwidth]{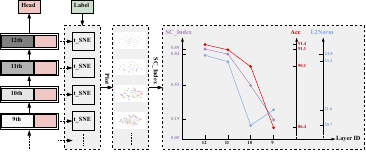}
    \caption{The outcomes of decreasing the dimensionality of RepAdapter\cite{repadapter} features from the $9^{th}$ to the $12^{th}$ layer on the Pets\cite{pets} Dataset, including the associated SC\_index for the features, model accuracy, and the L2Norm of the Adapter parameters in the respective layers.}
    \label{fig2}
\end{figure}
\textbf{t-SNE.} Given a set of $d$-dimensional input features $ X=\{x_1,x_2,...,x_n\}\in \mathbb{R}^{n\times d}$, t-SNE computes a set of $s$-dimensional embeddings for $X$, denoted as $ Y=\{y_1,y_2,...,y_n\}\in \mathbb{R}^{n\times s}$. Where $s\ll d$, usually 2 or 3 for visualization. t-SNE measures the similarity between $x_i$ and $x_j$ in the input $X$ using the following joint probability
\begin{equation}
\label{eq1}
p_{ij}=\frac{p_{i|j}+p_{j|i}}{2n}
\end{equation}
\begin{equation}
\label{eq2}
    p_{i|j}=\frac{\exp (-||x_i-x_j||^2/2 \sigma _i^2)}{\sum_{k\ne i}\exp(-||x_i-x_k||^2/2 \sigma _i^2)}
\end{equation}
The tuning parameters $\sigma_i$ are typically determined by utilizing a particular perplexity measure and a binary search strategy\cite{t-SNE}.

For randomly initialized $Y$, the similarity between $y_i$ and $y_j$ is defined as
\begin{equation}
\label{eq3}
q_{ij}=\frac{(1+||y_i-y_j||^2)^-1}{\sum_{k \ne l}(1+||y_k-y_l||^2)^-1}
\end{equation}Based on the above joint probabilities $P$ and $Q$ of $X$ and $Y$, t-SNE optimizes the Kullback-Leibler divergences between $P$ and $Q$ by gradient descent method, and the objective function is defined as
\begin{equation}
\label{eq4}
KL(P||Q)=\sum_{i \ne j} p_{ij}\log\frac{p_{ij}}{q_{ij}}
\end{equation}
\textbf{Silhouette Coefficient Index.} Given a set of clustering results $X=\{x_1,x_2,...,\\x_n\}$, for one point $x_i$, define $a(i)$ to be the average distance between the remaining points in its cluster and $x_i$, and $b(i)$ to be the average distance between all points in the nearest cluster and $x_i$. Then define the silhouette coefficient index of this clustering result as follows
\begin{equation}
\label{eq5}
\bar{s}=\frac{1}{n}\sum_{i=1}^n\frac{b(i)-a(i)}{\max(a(i),b(i))}
\end{equation}

\begin{figure*}[t]
    \centering
    \includegraphics[width=1\textwidth]{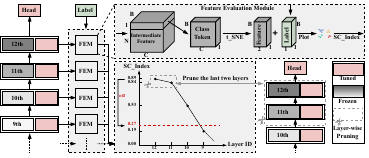}
    \caption{Overview of our SLS. By aggregating intermediate features output by each layer and calculating the SC\_Index of features, SLS dynamically determines the number of layers to prune based on the hyperparameter $\bm{\alpha}$.}
    \label{fig3}
\end{figure*}

\subsection{Layer-wise Pruning with Feature Perspective}
Previous studies \cite{specific_1, specific_2, specific_3} have shown that lower-level layers typically capture general features, while higher-level layers focus on specific features. Based on this understanding and the limitations of freezing pre-trained network parameters in the PETL model, we propose a hypothesis that when there is a significant difference between the downstream data distribution and the pre-training data distribution, the deeper layers of the model transferred by PETL will contain a large number of redundant parameters. Our objective is to dynamically identify and prune these redundant parameters. To ensure the effectiveness of the pruning process, the following key criteria must be met:
\begin{itemize}
\setlength{\itemsep}{4pt}
\setlength{\parsep}{4pt}
\setlength{\parskip}{4pt}
    \item Maintain the same storage parameter quantity as the PETL transfer method.
    \item Predict redundant parameters without requiring additional training.
\end{itemize}

\textbf{Pruning at the layer level does not increase the storage parameter quantity.} Consider a PETL-transferred ViT\cite{vit} with $N$ layers. Let the $i^{th}$ layer of the model be denoted as $L_i$ $(1 \leq i \leq N)$, and the pre-training parameters of each layer be represented as $\bm{W_P^i}$. For $K$ downstream datasets $D_1, D_2, ..., D_K$, we denote the newly introduced training parameters at each layer on dataset $D_j$ $(1 \leq j \leq K)$ (such as Adapter) as $\bm{W_A^{i,j}}$, the parameters of the head as $\bm{{W_H^j}}$, and the pruning layer index predicted by SLS as $Index_j$. Therefore, for the pruning method at the layer level, the storage parameter quantity $S$ across $K$ downstream datasets can be expressed as
\begin{equation}
\label{eq6}
S=\sum_{j=1}^{K} \sum_{i=1}^{Index_j} (\bm{W_P^i}+ \bm{W_A^{i,j}})+\sum_{j=1}^{K}\bm{W_H^j}
\end{equation}
Clearly, the number of pruned layers predicted by SLS will not exceed the range of network layers, i.e., $1 \leq Index_j \leq N$. Therefore
\begin{equation}
\label{eq7}
S\leq \sum_{j=1}^{K} \sum_{i=1}^{N} (\bm{W_P^i}+ \bm{W_A^{i,j}})+\sum_{j=1}^{K}\bm{W_H^j}=K\times\sum_{i=1}^{N} \bm{W_P^i}+\sum_{j=1}^{K} \sum_{i=1}^{N} \bm{W_A^{i,j}}+\sum_{j=1}^{K}\bm{W_H^j}
\end{equation}
In terms of storage parameters, the parameters of each layer in the pre-trained network, $\bm{W_P^i}$, are evidently reusable, hence
\begin{equation}
\label{eq8}
K\times\sum_{i=1}^{N} \bm{W_P^i}\Longleftrightarrow\sum_{i=1}^{N}\bm{W_P^i}
\end{equation}
\begin{equation}
\label{eq9}
S\leq \sum_{i=1}^{N} \bm{W_P^i} + \sum_{j=1}^{K} \sum_{i=1}^{N} \bm{W_A^{i,j}} + \sum_{j=1}^{K} \bm{W_H^j}
\end{equation}
This exactly equals the storage parameter quantity of the pre-trained model transferred using the PETL method across $K$ datasets. Therefore, a model pruned by SLS will not incur any additional storage overhead.

\textbf{Making pruning decisions based on the intermediate features from each layer.} We propose evaluating the layer features using the clustering degree SC\_Index\cite{SCIndex} of the reduced-dimensional features. This approach does not introduce additional supervised training. As shown in \cref{fig2}, under appropriate settings, there is a clear correlation between the classification accuracy of the current layer in the model and the SC\_Index of the reduced-dimensional features.

Given an input $X \in \mathbb{R} ^{B\times C\times H \times W}$, the ViT\cite{vit} model embeds it into a $d$-dimensional latent space through the patch embedding layer, resulting in $e_0 \in \mathbb{R}^{B\times M \times d}$, where $M$ is the token count in the ViT model. Subsequently, $e_0$ is concatenated with the cls\_token to serve as the model's input. We denote the embedding output of the $i^{th}$ layer in ViT as $e_i$ and the cls\_token as $x_i \in \mathbb{R}^{B\times d}$. Therefore, the forward process of the ViT model backbone can be expressed as
\begin{equation}
\label{eq10}
[x_i,e_i] =L_i([x_{i-1},e_{i-1}])\qquad i=1,2,...,N
\end{equation}

We propose the Feature Evaluation Module (FEM) to evaluate the feature from layer $L_i$. As shown in \cref{fig3}, the FEM takes the cls\_token $x_i$ from the output $[x_i,e_i]$ of layer $L_i$ to represent the current features. Then, it uses the t-SNE\cite{t-SNE} algorithm to reduce $x_i$ to $x_i' \in \mathbb{R}^{B\times2}$. Subsequently, by combining the label $\in \mathbb{R}^{B\times1}$ corresponding to the current input, a clustering result $C$ with $p$ classes, where $p$ is the number of classes in the current dataset, is obtained. Next, the values $a(i)$ and $b(i)$ corresponding to $C$ are calculated, and finally, the evaluation of the current layer's features, $SC\_index_i$, is determined using \cref{eq5}.

For a model with $N$ layers, let $\bm{\alpha}$ be a hyperparameter that controls the degree of pruning by SLS. The threshold $T$ for the number of pruned layers on the current dataset is defined as
\begin{equation}
\label{eq11}
T=\bm{\alpha} \times SC\_Index_N
\end{equation}

During the pruning process of the model, we traverse from the highest layer downwards. When the evaluation of the $i^{th}$ layer features, $SC\_Index_i$, is less than $T$, we stop the traversal loop and prune layers from $i+2$ to $N$. The motivation behind this design is that when the evaluation of the $i^{th}$ layer features is below a certain threshold compared to the top layer features evaluation, the classification head is no longer able to distinguish the current features effectively. Therefore, the output features of the ${i+1}^{th}$ layer are the lowest layer that can be well distinguished by the classification head. Pruning the layers from $i+1$ downwards significantly impacts the model's performance.
\section{Experiment}
\subsection{Dataset and Metric}
\textbf{Image Classification.} We utilized the VTAB-1K\cite{vtab} benchmark as our experimental dataset, a transfer benchmark composed of 19 datasets that can be categorized into Natural, Specialized, and Structured groups according to their data types. The Natural category is primarily comprised of natural image datasets, such as CIFAR\cite{cifar} and Flower102\cite{flower}. The Specialized category consists mainly of professional image datasets, such as Retinopathy\cite{retinopathy}. The Structured class consists of structured image data such as dspr-loc\cite{dspr-loc}, which can be considered as cross-domain data to ImageNet\cite{imagenet} dataset. According to previous studies\cite{vpt,convpass,repadapter}, the model was trained using the train-val split in the VTAB-1k benchmark and reported top1-accuracy on the test split. Additionally, we measure the efficiency of both the current PETL model and the pruned model using throughput. The throughput result is reported as the average of 19 datasets on an RTX-3090 platform with batch\_size $=$ 64. 

\noindent\textbf{Object Detection.} We use MS COCO\cite{coco} as the dataset for object detection experiments, which contains about 118K images with bounding boxes and
instance segmentation annotations and covers 80 object categories in total. We report $mAP$ on its validation set.

\noindent\textbf{Image Segmentation.} We use ADE20K\cite{ade20k} as the dataset for our experiments. ADE20K poses a significant challenge for semantic segmentation tasks, comprising 20,000 images for training and 2,000 images for validation across 150 diverse categories. We report $mIoU$ on its validation set.

\begin{table*}[t]
\caption{Comparison of pruned RepAdapter\cite{repadapter} and Convpass\cite{convpass} with their unpruned model and other PETL methods on VTAB-1k\cite{vtab} benchmark. Pruned Conv. and Pruned Rep. refer to the Convpass and RepAdapter models pruned by our SLS algorithm. ViT-B/16 pre-trained on ImageNet-21k is used as the vision model of all methods.}
\setlength\tabcolsep{1.5pt}
\renewcommand{\arraystretch}{1.2}
\centering
\scalebox{0.68}{
\begin{tabular}{@{}lccc|ccccccc|cccc|cccccccc@{}}
\toprule
\multicolumn{1}{c}{} &
   &
  \multicolumn{1}{l}{} &
   &
  \multicolumn{7}{c|}{\textbf{Natural}} &
  \multicolumn{4}{c|}{\textbf{Specialized}} &
  \multicolumn{8}{c}{\textbf{Structured}} \\
Model &
\rotatebox{90}{Params (M)} &
  \rotatebox{90}{Throughput} &
  \rotatebox{90}{Avg Accuracy} &
  \rotatebox{90}{Cifar100} &
  \rotatebox{90}{Caltech101} &
  \rotatebox{90}{DTD} &
  \rotatebox{90}{Flower102} &
  \rotatebox{90}{Pets} &
  \rotatebox{90}{SVHN} &
  \rotatebox{90}{Sun397} &
  \rotatebox{90}{Camelyon} &
  \rotatebox{90}{EuroSAT} &
  \rotatebox{90}{Resisc45} &
  \rotatebox{90}{Retinopathy} &
  \rotatebox{90}{Clevr-Count} &
  \rotatebox{90}{Clevr-Dist} &
  \rotatebox{90}{DMLab} &
  \rotatebox{90}{KITTI-Dist} &
  \rotatebox{90}{dspr-Loc} &
  \rotatebox{90}{dspr-Ori} &
  \rotatebox{90}{sNORB-Azim} &
  \rotatebox{90}{sNORB-ele}  \\ \midrule
Full-Tuning\cite{vpt} &
  85.8 &
  1$\times$ &
  68.9 &
  68.9 &
  87.7 &
  64.3 &
  97.2 &
  86.9 &
  87.4 &
  38.8 &
  79.9 &
  95.7 &
  84.2 &
  73.9 &
  56.3 &
  58.6 &
  41.7 &
  65.5 &
  57.5 &
  46.7 &
  25.7 &
  29.1 \\
Linear probe\cite{vpt} &
  0.04 &
  1$\times$ &
  57.6 &
  64.4 &
  85.0 &
  63.2 &
  97.0 &
  86.3 &
  36.6 &
  51.0 &
  78.5 &
  87.5 &
  68.5 &
  74.0 &
  34.3 &
  30.6 &
  33.2 &
  55.4 &
  12.5 &
  20.0 &
  9.6 &
  19.2 \\ \midrule
VPT\cite{vpt} &
  0.64 &
  0.71$\times$ &
  72.0 &
  78.8 &
  90.8 &
  65.8 &
  98.0 &
  88.3 &
  78.1 &
  49.6 &
  81.8 &
  96.1 &
  83.4 &
  68.4 &
  68.5 &
  60.0 &
  46.5 &
  72.8 &
  73.6 &
  47.9 &
  32.9 &
  37.8 \\
LoRA\cite{lora} &
  0.29 &
  1$\times$ &
  74.5 &
  67.1 &
  91.4 &
  69.4 &
  98.8 &
  90.4 &
  85.3 &
  54.0 &
  84.9 &
  95.3 &
  84.4 &
  73.6 &
  82.9 &
  69.2 &
  49.8 &
  78.5 &
  75.7 &
  47.1 &
  31.0 &
  44.0 \\
AdaptFormer\cite{adaptformer} &
  0.16 &
  0.69$\times$ &
  74.7 &
  70.8 &
  91.2 &
  70.5 &
  99.1 &
  90.9 &
  86.6 &
  54.8 &
  83.0 &
  95.8 &
  84.4 &
  76.3 &
  81.9 &
  64.3 &
  49.3 &
  80.3 &
  76.3 &
  45.7 &
  31.7 &
  41.1 \\ \midrule
Convpass\cite{convpass} &
  0.33 &
  0.85$\times$ &
  76.7 &
  72.2 &
  90.7 &
  72.3 &
  99.2 &
  90.8 &
  91.3 &
  55.0 &
  86.0 &
  95.9 &
  85.6 &
  76.1 &
  81.7 &
  67.7 &
  51.3 &
  82.6 &
  84.8 &
  53.6 &
  35.2 &
  43.3 \\
\rowcolor[HTML]{E6E6E6} 
Pruned Conv. &
  0.26 &
  1.16$\times$ &
  75.4 &
  71.6 &
  88.8 &
  72.3 &
  98.3 &
  84.7 &
  88.9 &
  55.0 &
  85.1 &
  94.7 &
  82.0 &
  76.1 &
  81.5 &
  68.2 &
  49.8 &
  81.0 &
  82.6 &
  53.1 &
  35.3 &
  43.3 \\
RepAdapter\cite{repadapter} &
  0.24 &
  1$\times$ &
  75.8 &
  72.4 &
  91.1 &
  69.5 &
  99.2 &
  91.4 &
  89.9 &
  55.4 &
  85.4 &
  95.9 &
  84.8 &
  75.3 &
  81.1 &
  68.9 &
  50.6 &
  81.7 &
  79.5 &
  47.1 &
  36.3 &
  41.0 \\
\rowcolor[HTML]{E6E6E6} 
Pruned Rep. &
  0.18 &
  1.42$\times$ &
  74.8 &
  71.4 &
  87.3 &
  68.1 &
  96.0 &
  89.9 &
  89.3 &
  53.4 &
  85.0 &
  95.3 &
  81.9 &
  75.2 &
  80.9 &
  69.8 &
  50.5 &
  80.7 &
  80.5 &
  47.1 &
  35.7 &
  41.0 \\ \bottomrule
\end{tabular}}
\label{table1}
\end{table*}
\subsection{Baseline}
\textbf{PETL Method.} Our task is to make the model more efficient by using simple pruning operations and maintaining almost perfect accuracy in PETL.  We aim to verify the applicability of our method in PETL by conducting experiments with the add-based sota PETL method Convpass\cite{convpass} and the reparameterization-based sota PETL method RepAdapter\cite{repadapter} as the baseline models. 

\noindent\textbf{Pruning Method.} To confirm that our pruning method is better than traditional structural pruning in PETL, we chose DepGraph\cite{depgraph} as our basis for pruning. This includes DepGraph-Hessian based on gradient and DepGraph-L2 Norm based on magnitude.

\subsection{Training Details}
\textbf{Hyperparameters for Retraining.} For the hyperparameters of the retraining process, we directly utilized the same hyperparameters as the training process and employed the 1-CLR\cite{1-clr} strategy as the default setting. Please refer to the Appendix for more information.

\noindent\textbf{Hyperparameters for SLS.} In this study, we set the hyperparameter $\bm{\alpha}$ to 0.3, which is an empirical parameter. We found that this setting balances model accuracy and speed in the current experiment. For the selection of t-SNE\cite{t-SNE} hyperparameters, we chose to project high-dimensional features into 2 dimensions (n\_components=2) based on experimental results and \cref{fig2}. This demonstrates a high correlation between clustering metrics based on only 2-dimensional features computation and model accuracy. Given the number of categories in the VTAB-1K\cite{vtab} benchmark, we set the perplexity of t-SNE to 30, the learning rate to 50, and n-iter to 1000 to ensure appropriate clustering of high-dimensional features within a reasonable range. The remaining hyperparameters utilize the default parameters of the sklearn.TSNE class.

\noindent\textbf{Other Details.} For experiments conducted on Swin\cite{swin}, the model layers are the only ones pruned due to different feature dimensions at various stages, which complicates the assessment of layer importance. We retain the Patch Merging module to align feature dimensions. In segmentation experiments, as the SETR-PUP\cite{setr}'s ViT structure contains auxiliary segmentation heads, to preserve the model's multi-scale reasoning capability, we pruned only the ViT layer structure during the pruning process while retaining the auxiliary heads.

\begin{table*}[t]
\caption{Comparison of SLS and DepGraph\cite{depgraph} on VTAB-1k\cite{vtab} benchmark with the same pruned parameters. Stored Params is the average value on 19 datasets. The average accuracy of the natural, specialized, and structured groups are denoted as "Nat.", "Spe.", and "Str.", respectively.}
\centering
\scalebox{0.87}{
\begin{tabular}{cccccccc}
\toprule
PETL Model & Pruning Method               & \multicolumn{1}{l}{Stored Params (M)} & Throughput & Avg Acc. & Nat. & Spe. & Str. \\ \hline
           & -  & 0.33                              & 0.85$\times$      & 76.7     & 81.7 & 85.9 & 62.5 \\
           & DepGraph-L2Norm\cite{depgraph}  & 66.6                              & 1.004$\times$      & 72.4     & 76.5 & 79.8 & 61.0 \\
           & DepGraph-Hessian\cite{depgraph} & 66.6                              & 1.004$\times$      & 75.0     & 79.7 & 83.4 & 61.9 \\
\multirow{-4}{*}{Convpass\cite{convpass}} &
  \cellcolor[HTML]{E6E6E6}SLS &
  \cellcolor[HTML]{E6E6E6}0.26 &
  \cellcolor[HTML]{E6E6E6}1.16$\times$ &
  \cellcolor[HTML]{E6E6E6}75.4 &
  \cellcolor[HTML]{E6E6E6}79.9 &
  \cellcolor[HTML]{E6E6E6}84.5 &
  \cellcolor[HTML]{E6E6E6}61.9 \\ \hline
           & -              & 0.24                              & 1$\times$      & 75.8     & 81.2 & 85.4 & 60.8 \\
           & DepGraph-L2Norm\cite{depgraph}              & 56.4                              & 1.18$\times$      & 71.0     & 75.8 & 79.3 & 57.8 \\
           & DepGraph-Hessian\cite{depgraph}             & 56.4                              & 1.18$\times$      & 74.0     & 79.3 & 83.2 & 59.6 \\
\multirow{-4}{*}{RepAdapter\cite{repadapter}} &
  \cellcolor[HTML]{E6E6E6}SLS &
  \cellcolor[HTML]{E6E6E6}0.18 &
  \cellcolor[HTML]{E6E6E6}1.42$\times$ &
  \cellcolor[HTML]{E6E6E6}74.8 &
  \cellcolor[HTML]{E6E6E6}79.3 &
  \cellcolor[HTML]{E6E6E6}84.4 &
  \cellcolor[HTML]{E6E6E6}60.8 \\  \bottomrule
\end{tabular}}
\label{table2}
\end{table*}
\begin{figure}[t]
    \centering
    \includegraphics[scale=0.8]{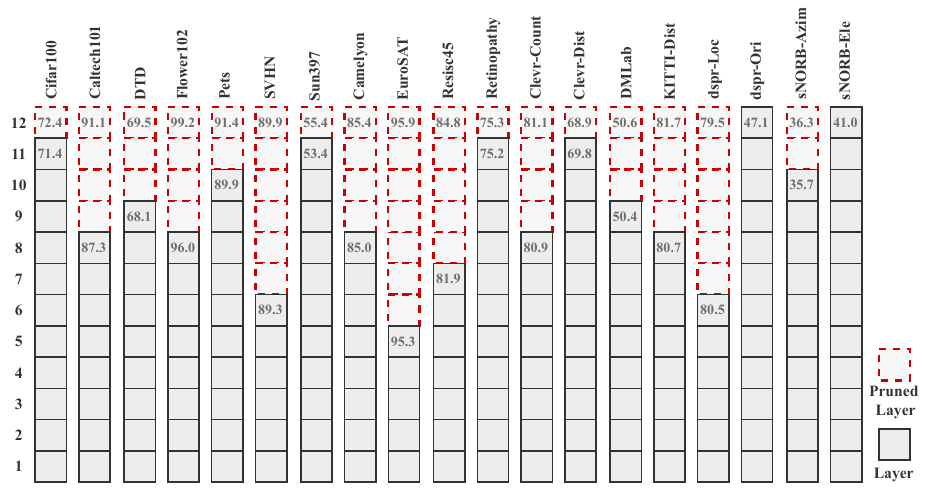}
    \caption{The pruning outcomes of RepAdapter\cite{repadapter} on VTAB-1K\cite{vtab} and the corresponding precision of the model before and after pruning.}
    \label{fig4}
\end{figure}
\subsection{Main Results}
\textbf{Comparison with the Sota PETL.} We compared the performance of the two PETL methods after pruning with other PETL methods on the VTAB-1k\cite{vtab} benchmark, as shown in \cref{table1}. The results illustrate that even with some loss of accuracy, the two pruned PETL methods still outperformed various PETL methods. In addition, the pruned Convpass and RepAdapter significantly enhance the model's throughput and decrease the number of updated parameters when compared to their unpruned counterparts, all while staying within an acceptable accuracy degradation range. For the dspr-Loc\cite{dspr-loc} dataset, the accuracy of the RepAdapter method increased by 1.0\% after pruning, verifying the existence of data redundancy in the pre-trained ImageNet\cite{imagenet} model. \cref{fig4} shows the number of pruning layers for RepAdapter on different datasets of VTAB-1K, visually illustrating the varying redundancy levels of pre-trained models on different datasets.

\noindent\textbf{Comparison with the Mainstream Pruning.} \cref{table2} displays our SLS performance compared to the mainstream pruning method, DepGraph\cite{depgraph}, on VTAB-1k. SLS performs better than DepGraph in terms of model accuracy, whether the results are based on Convpass or RepAdapter. This is because the trainable parameters in the model need to adapt to the pruned new structure during the retraining process. SLS requires fewer stored parameters compared to both the magnitude-based L2Norm and the gradient-based Hessian. This is because pruning methods prune models at different rates for various datasets, resulting in different structures to be stored. Storing these structures is costly, whereas our method only requires the index of the pruning layers and a few adapter parameters. As a result, SLS greatly reduces the number of stored parameters. Additionally, most existing pruning methods attempt to reduce the size of a wide network to a narrow one, whereas our approach aims to prune a deep network into a shallow one, resulting in better model throughput compared to traditional pruning methods.

\noindent\textbf{Ablation Studies.} To gain more understanding of SLS, we conducted many ablation experiments for both pruning and retraining strategies.  The results are shown in \cref{table3} and \cref{table4}. \cref{table3} focuses on the impact of the pruning strategy, with iterative pruning referring to progressively pruning layers until our algorithm's target number is reached. After each layer is pruned, retraining the pruned model in an attempt to help the model regain some of its lost accuracy. One-shot pruning involves pruning the model to the target number of layers and then performing retraining, so the pruning process is only performed once retrained, enhancing pruning efficiency. For both Convpass\cite{convpass} and RepAdapter\cite{repadapter}, we can observe that iterative pruning is better than direct pruning solely in terms of accuracy. Iterative pruning requires more training than direct pruning because it requires multiple retraining sessions. Even after several sessions, the experimental results revealed that iterative pruning has limited improvement when compared to direct pruning, with only a 0.1\% improvement on the RepAdapter. This differs from the results for conventional models, and we hypothesize that this is due to the PETL model using only a small number of parameters for learning. 

\begin{table}[t]
\caption{Ablation study for pruning strategies on VTAB-1K\cite{vtab}. We conducted experiments on both One-shot and Iterative strategies, with the gray background strategy set as the default setting.}
\centering
\setlength{\tabcolsep}{5pt}
\begin{tabular}{cccccc}
\toprule
Method                    & Strategy  & Avg Accuracy      & Natural & Specialized & Structured \\ \hline
\multirow{2}{*}{Convpass\cite{convpass}} & \cellcolor[HTML]{E6E6E6} One-shot  & \cellcolor[HTML]{E6E6E6} 75.4 & \cellcolor[HTML]{E6E6E6} 79.9 & \cellcolor[HTML]{E6E6E6} 84.5 & \cellcolor[HTML]{E6E6E6} 61.9 \\ 
                          & Iterative & \cellcolor[HTML]{FFFFFF} 75.7 &   \cellcolor[HTML]{FFFFFF} 80.0 &   \cellcolor[HTML]{FFFFFF} 84.8 & 
  \cellcolor[HTML]{FFFFFF} 62.2 \\ \hline
\multirow{2}{*}{RepAdapter\cite{repadapter}}  & \cellcolor[HTML]{E6E6E6} One-shot  & \cellcolor[HTML]{E6E6E6} 74.8 &  \cellcolor[HTML]{E6E6E6} 79.3 &  \cellcolor[HTML]{E6E6E6} 84.4 &  \cellcolor[HTML]{E6E6E6} 60.8 \\ 
& Iterative &  \cellcolor[HTML]{FFFFFF}  74.9 &  \cellcolor[HTML]{FFFFFF}  79.7 &  \cellcolor[HTML]{FFFFFF}  85.0 &  \cellcolor[HTML]{FFFFFF}  59.9\\ \bottomrule
\end{tabular}
\label{table3}
\end{table}

\begin{table}[t]
\caption{Ablation study for retraining strategies for Convpass on VTAB-1K\cite{vtab}. We conducted experiments on three common retraining strategies: Training From Scratch, Fine-Tuning, and Cyclic Learning Rate Restarting.}
\centering
\setlength{\tabcolsep}{5pt}
\begin{tabular}{cccccc}
   \toprule
   Strategy & Avg Accuracy & Natural & Specialized & Structured\\
   \midrule
    TFS & 74.4 & 78.4 & 83.4 & 61.3\\
  FT  & 73.5 & 77.6 & 82.2 & 60.7\\
  \rowcolor[HTML]{E6E6E6} 1-CLR\cite{1-clr} & 75.4 & 79.9 & 84.5 & 61.9\\
   \bottomrule
\end{tabular}
\label{table4}
\end{table}

In \cref{table4}, we show the impact of different retraining strategies. Where Cyclic Learning Rate Restarting (CLR) indicates that the learning rate during retraining varies cyclically from maximum to minimum based on the cosine function, following the previous work\cite{1-clr}, we use 1-CLR for training. Training From Scratch (TFS) means that the original parameters are not retained for the pruned model structure, but the parameters are reinitialized for training. Fine-Tuning (FT) means that the model is pruned and continues to be trained using only the current learning rate (or setting a small learning rate). We can see that 1-CLR performs significantly better than the remaining two strategies. The performance of TFS decreased by 1\%, which is perhaps caused by the fact that the PETL method is quite sensitive to parameter initialization. The 1-CLR strategies Adapter is used for prior information with current data, while the TFS with random initialization needs to be relearned. The outcome of FT could be attributed to the absence of a high learning rate utilized in the retraining phase\cite{retrain}.

\begin{table}[t]
\caption{The pruning results of Convpass and RepAdapter based on the Swin-B backbone on VTAB-1K. We employ one-shot as the pruning strategy and 1-CLR as the retraining strategy.}
\centering
\setlength{\tabcolsep}{5pt}
\scalebox{0.93}{
\begin{tabular}{cccccc}
   \toprule
    Method & Throughput & Avg Accuracy & Natural & Specialized & Structured\\
   \midrule
  Full-Tuning\cite{vpt} & 1$\times$ & 75.0 & 79.2 & 86.2 & 59.7\\
   Linear probe\cite{vpt} & 1$\times$ & 62.6 & 73.5 & 80.8 & 33.5 \\
  \midrule Convpass\cite{convpass} & 0.80$\times$ & 78.4 & 82.9 & 87.8 &64.5\\
   \rowcolor[HTML]{E6E6E6} Pruned ConvPass & 0.90$\times$ & 77.5 & 81.0 & 86.6 &64.8\\
  RepAdapter\cite{repadapter} & 1$\times$ & 77.2 & 82.6 & 87.5 &61.4\\
   \rowcolor[HTML]{E6E6E6} Pruned RepAdapter & 1.16$\times$ & 76.7 & 81.5 & 86.6 &62.0\\
   \bottomrule
\end{tabular}}
\label{table5}
\end{table}

\subsection{Results of More Model Structure}
Unlike ViT\cite{vit}, Swin\cite{swin} borrows the inductive bias from the convolutional model, and outputs features of different dimensions at different stages, which may affect the effectiveness of layer-wise pruning, as compared to ViT, where the difference in features between layers may be greater. To explore this effect, we conducted one-shot pruning experiments on Swin using Convpass and RepAdapter on VTAB-1k\cite{vtab}. The experimental results are shown in \cref{table5}.
We observe that both the Convpass and RepAdapter perform better in the structured category after pruning compared to the unpruned model, demonstrating the significant redundancy structure of the model when encountering data distributions that differ significantly from the ImageNet\cite{vtab}.

\begin{table}[t]
\caption{Results of detection task on COCO\cite{coco}. The throughput is evaluated on an NVIDIA A800 platform with batch\_size=1.}
\centering
\setlength{\tabcolsep}{6.5pt}
\begin{tabular}{ccccc}
\toprule
Method                       & Throughput & mAP  & mAP$_{50}$ & mAP$_{75}$ \\ \midrule
Head only                    & 1$\times$     & 28.8 & 49.8    & 29.0    \\
RepAdapter\cite{repadapter}                   & 1$\times$     & 32.4 & 53.2    & 33.6    \\ \midrule
\rowcolor[HTML]{E6E6E6}
Pruned RepAdapter (Prune 2 Layers) & 1.22$\times$      &33.2      & 53.1        &  34.6     \\
\rowcolor[HTML]{E6E6E6}
Pruned RepAdapter (Prune 4 Layers) & 1.43$\times$       &30.7      & 49.5        & 32.2        \\ \bottomrule
\end{tabular}
\label{detection}
\end{table}

\begin{table}[t]
\caption{Results of segmentation task on ADE20K\cite{ade20k}. The throughput is evaluated on an RTX A5000 platform with batch\_size=1. "mIoU-SS" and "mIoU-MS" represent the results of single-scale and multi-scale predictions, respectively.}
\centering
\setlength{\tabcolsep}{8pt}
\begin{tabular}{cccc}
\toprule
Method                  & Throughput & mIoU-SS & mIoU-MS \\ \midrule
Head only\cite{setr}              & 1$\times$   & 35.12   & 37.46         \\
RepAdapter\cite{repadapter}             & 1$\times$   & 41.40   & 43.42         \\ \midrule
\rowcolor[HTML]{E6E6E6} 
Pruned RepAdapter (Prune 6 Layers) & 1.26$\times$   & 40.59   & 42.36       \\
\rowcolor[HTML]{E6E6E6} Pruned RepAdapter (Prune 8 Layers) & 1.38$\times$   & 40.05   & 41.82       \\ \bottomrule
\end{tabular}
\label{segmentation}
\end{table}

\subsection{Results of More Vision Tasks}
To validate the generalizability of SLS beyond image classification tasks, we applied SLS to object detection and image segmentation tasks.

\noindent\textbf{Object Detection.}
We conducted experiments on the COCO\cite{coco} dataset using RetinaNet\cite{retinanet} with a Swin-Tiny\cite{swin} backbone. Initially, we integrated RepAdapter into the pre-trained Swin-Tiny backbone, freezing all parameters except the detection head and RepAdapter, and trained this subset of parameters. Subsequently, we pruned the Swin-Tiny part of RetinaNet using SLS and followed a 1-CLR\cite{1-clr} strategy to retrain the pruned model. As shown in \cref{detection}, two sets of experiments with different hyperparameters($\alpha$) were performed, leading to two distinct pruning outcomes (pruning 2 and 4 layers). Both pruned models significantly improve the model’s speed. The model with two pruned layers even outperforms the original model.

\noindent\textbf{Image Segmentation.}
We conducted experiments on the ADE20K\cite{ade20k} dataset using the SETR-PUP\cite{setr}. Specifically, SETR-PUP is a segmentation model with a ViT\cite{vit} backbone and a convolutional network as the segmentation head. Initially, we integrated RepAdapter into the pre-trained ViT backbone on ImageNet21k\cite{imagenet} within SETR-PUP, freezing all parameters except RepAdapter\cite{repadapter} and the segmentation head, and trained this subset of parameters on ADE20K. Subsequently, we pruned the ViT part of SETR-PUP using SLS and followed a 1-CLR\cite{1-clr} strategy to retrain the pruned model. As shown in \cref{segmentation}, two sets of experiments with different $\alpha$ were performed, leading to two distinct pruning outcomes (pruning 6 and 8 layers). Notably, both sets of experiments showcased significant enhancements in model speedup while maintaining acceptable performance drop. 

\section{Limitation}
While pruning an entire layer like our SLS directly offers many advantages, such as low model inference latency and low model storage parameters, it inevitably results in the loss of pre-trained information, even if it has little impact on the current data. Traditional structural pruning methods, on the other hand, can selectively prune parameters. However, they significantly increase the number of stored parameters under different pruning rates and input data conditions. Therefore, maintaining storage parameter quantity while enhancing granularity for PETL method pruning presents a challenging issue.

\section{Conclusion}
In this paper, we present a straightforward but powerful pruning method called SLS. SLS makes pruning decisions through a novel feature perspective, avoiding the introduction of additional training overhead. By pruning the deep network to the shallow network, SLS significantly enhances the inference speed of the model without high accuracy losses. Additionally, SLS considerably reduces the storage overhead of commonly used pruning methods. This is achieved by directly pruning the entire layer structure and only requiring the index of the specific layers to be pruned for various downstream data. To verify the applicability of SLS to the PETL method, we employed it in two mainstream PETL methods and conducted comprehensive experiments on 21 datasets for three vision tasks. To evaluate the benefits of SLS and its superiority over the mainstream pruning methods, we conducted a comparative analysis with the mainstream pruning method. The experimental results demonstrate the superiority of SLS in terms of accuracy, efficiency, and storage.\\

\noindent\textbf{Acknowledgements.} This work was supported by Key Research and Development projects in Shaanxi Province (No. 2023-YBNY-121), Key Research and Development projects in Shaanxi Province (No. 2023-YBNY-080), Xi’an Science and Technology Plan Project (No. 22NYYF013), and the Xianyang Key Project of Research and Development Plan (No. L2022ZDYFSF050).

\clearpage
\title{Appendix}
\author{}
\institute{}
\maketitle
\setcounter{section}{0}
\renewcommand{\thesection}{\arabic{section}}

\section{More Experiment Details}
\textbf{Pre-trained Backbones.} We conducted experiments utilizing the ViT-B/16\cite{vit} and Swin-B\cite{swin} models built with the Timm\cite{timm} library, both of which were pre-trained on ImageNet21K\cite{imagenet}.

\noindent\textbf{Code Implementation.} We employed PyTorch to conduct our primary experiments on an NVIDIA RTX-A5000 GPU and evaluated the model throughput using an NVIDIA RTX3090 GPU. The pytorch-like pseudo-code for calculating model throughput is as follows \cref{throughput}.

\noindent\textbf{Data Augmentation.} We resized the VTAB-1k\cite{vtab} images to 224$\times$224 and then normalized them using the mean and variance of the ImageNet, following\cite{convpass,repadapter}.

\noindent\textbf{Training Details.} Optimizer and hyper-parameters are shown in \cref{hyper-parameters}.

\section{More Results}
\begin{figure*}[t]
    \centering
    \includegraphics[scale=0.8]{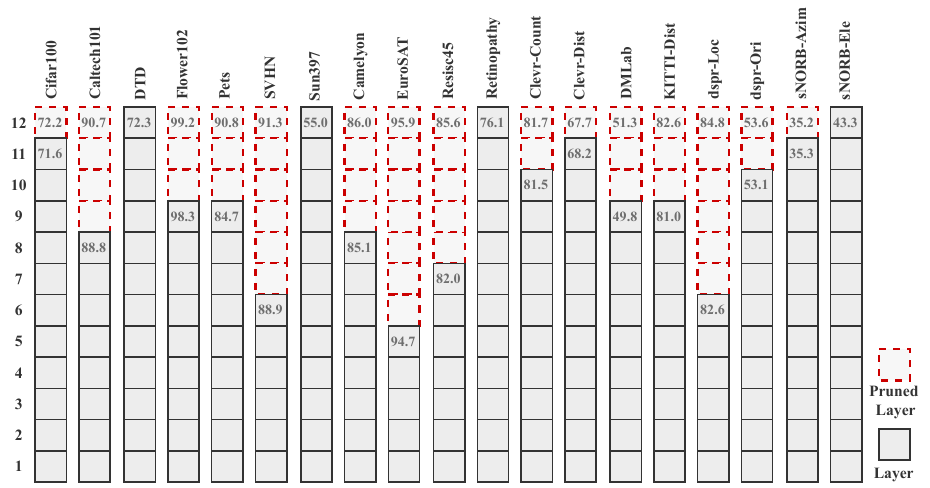}
    \caption{The pruning outcomes of Convpass on VTAB-1K and the corresponding precision of the model before and after pruning.}
    \label{convpass_result}
\end{figure*}

\begin{figure*}
    \centering
    \includegraphics[scale=0.18]{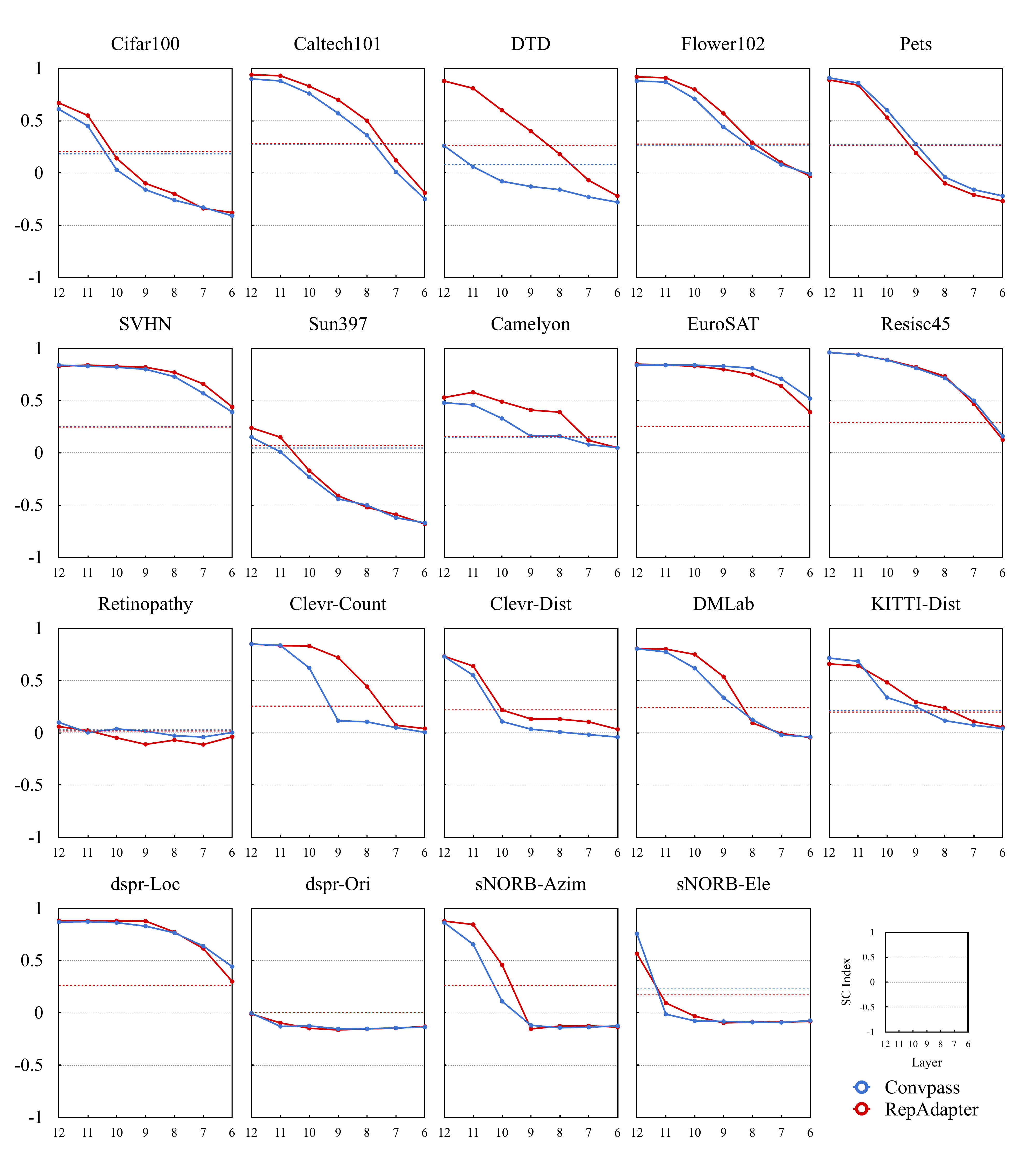}
    \caption{The relationship between the number of layers for Convpass and RepAdapter and SC\_index values on VTAB-1k.}
    \label{sc_index result}
\end{figure*}

\textbf{Detailed Results for Convpass.} Pruning results for the Convpass\cite{convpass} on VTAB-1k are depicted in \cref{convpass_result}, along with its corresponding model accuracy. The figure illustrates that an acceptable model accuracy can be maintained even with a significant number of pruned model structures for certain datasets, particularly SVHN, EuroSAT, and dspr-Loc. This finding aligns with the trend of SC\_index\cite{SCIndex} depicted in \cref{sc_index result}, which indicates a gradual decline in SC\_index for both the Convpass and RepAdapter\cite{repadapter} on these datasets as the number of layers decreases.

\definecolor{commentcolor}{RGB}{110,154,155}   
\newcommand{\PyComment}[1]{\ttfamily\textcolor{commentcolor}{\# #1}}  
\newcommand{\PyCode}[1]{\ttfamily\textcolor{black}{#1}} 

\begin{algorithm}[ht]
\renewcommand{\thealgocf}{1}
\SetAlgoLined
    \PyComment{Define an input tensor}\\
    \PyComment{x.shape = [B C H W]}\\
    \PyCode{x = torch.randn(64, 3, 224, 224)}\\
    \PyCode{batch\_size = x.shape[0]}\\
    \PyComment{Waits for all kernels to complete}\\
    \PyCode{torch.cuda.synchronize()}\\
    \PyCode{tic1 = time.time()}\\
    \PyComment{Inference begin}\\
    \PyCode{for i in range(100):}\\
    \Indp
        \PyCode{model(x)}\\
    \Indm
    \PyCode{torch.cuda.synchronize()}\\
    \PyCode{tic2 = time.time()}\\
    \PyCode{time = tic2 - tic1}\\
    \PyComment{Throughput calculate}\\
    \PyCode{throughput = 100 $\times$ batch\_size/time}\\
\caption{Calculating Model Throughput}
\label{throughput}
\end{algorithm}
\noindent\textbf{Analysis of Silhouette Coefficient Index.} Variation statistics on the relationship between the number of layers for Convpass and RepAdapter and SC\_index values across all datasets are displayed in \cref{sc_index result}. The horizontal axis represents the number of layers within the model, while the vertical axis represents the SC\_index computation from the current layer output features after t\_SNE\cite{t-SNE} dimensionality reduction. In each sub-figure, dashed lines represent the pruning thresholds of the two methods with respect to the current data, and pruning ceases at the point in which the number of layers falls below this threshold. The SC\_index variation significantly differs for various datasets. Datasets with high initial values and a gradual decrease in the number of layers, for instance, SVHN and EuroSAT, demonstrate significantly better pruning performance compared to those with lower starting values, like Sun397, or datasets with a rapid decline such as Pets.

\noindent\textbf{Results of Different DepGraph Settings}
Experiments were conducted to determine the effectiveness of the DepGraph\cite{depgraph} pruning method in pruning Adapter parameters and the retraining strategy. As DepGraph is a group pruning method, excluding certain modules could potentially affect the process of building the dependency graph and, in turn, impact the pruning results. The experimental outcomes are presented in \cref{depgraph-setting}. We observed a decrease in DepGraph's performance after pruning the adapter when compared to ignoring it. This outcome aligns with expectations, as pruning the adapter reduces the number of tunable parameters during the retraining process. The DepGraph with full retraining setting performed lower than the setting of using only the Adapter and Head, which may be due to overfitting caused by excessive parameters.
\begin{table}[t]
\caption{Comparison of different DepGraph\cite{depgraph} settings and SLS on
VTAB-1k\cite{vtab}. Pruning Adapter indicates that Depgraph prunes the Adapter parameters in the model while Ignoring Adapter means pruning only the pre-trained parameters in the model. $\dagger$To maintain training stability, we reduced the learning rate in the Full retraining setting.}
\centering
\setlength{\tabcolsep}{5pt}
\begin{tabular}{cccccc}
\toprule
Pruning Method                       & Retraining Setting & Avg Acc. & Nat. & Spe. & Str. \\ \midrule
DepGraph (Pruning Adapter)  & Adapter+Head       & 74.5         & 79.3    & 82.3        & 61.8       \\
DepGraph (Ignoring Adapter) & Full$^\dagger$               & 74.0         & 79.1    & 83.0        & 59.9       \\
DepGraph (Ignoring Adapter) & Adapter+Head       & 75.0         & 79.7    & 83.4        & 61.9       \\ \midrule
\rowcolor[HTML]{E6E6E6}SLS                         & Adapter+Head       & 75.4         & 79.9    & 84.5        & 61.9       \\ \bottomrule
\end{tabular}
\label{depgraph-setting}
\end{table}

\begin{table*}[t]
\centering
\setlength{\tabcolsep}{5pt}
\caption{Optimizer and hyper-parameters for 1-CLR\cite{1-clr} and TFS retraining strategies.}
\scalebox{0.9}{
\begin{tabular}{@{}cclllll@{}}
\toprule
optimizer & batch size & learning rate            & weight decay             & epochs               & lr decay                   & warm-up epochs       \\ \midrule
AdamW\cite{adamw}     & 64         & \multicolumn{1}{c}{1e-3} & \multicolumn{1}{c}{1e-4} & \multicolumn{1}{c}{100} & \multicolumn{1}{c}{cosine} & \multicolumn{1}{c}{10} \\ \bottomrule
\end{tabular}}
\label{hyper-parameters}
\end{table*}


\clearpage  

%
%

\bibliographystyle{splncs04}
\bibliography{main}

\end{document}